\documentclass[journal]{IEEEtran}

\usepackage{amsmath,amssymb,amsfonts}
\usepackage{algorithmic}
\usepackage{algorithm}
\usepackage{array}
\usepackage{textcomp}
\usepackage{stfloats}
\usepackage{url}
\usepackage{verbatim}
\usepackage{graphicx,color}
\usepackage{cite}
\usepackage{subcaption}

\begin{document}

\title{Multi-AUV Marine Life Tracking with Single Hydrophone Payloads via a Hidden Markov Model Equipped Particle Filter}

\author{Christopher Herrera, Kehlani Fay, Christopher Clark, Alberto Soto, Christopher Lowe, Mario Espinoza
\thanks{Christopher Herrera, Kehlani Fay, Christopher Clark, and Alberto Soto were with Harvey Mudd College, Claremont, CA 91711 USA (e-mail: ciherrera@g.hmc.edu, kfay@g.hmc.edu, clark@g.hmc.edu, alsoto@g.hmc.edu)}%
\thanks{Christopher G. Lowe is with California State University Long Beach, Long Beach, CA 90840 USA (e-mail: chris.lowe@csulb.edu)}%
\thanks{Mario Espinoza is with University of Costa Rica, San Jose, Costa Rica (e-mail: mario.espinoza\_m@ucr.ac.cr)}
}


 
\maketitle

\begin{abstract}
Researchers tag and track marine animals to study migration patterns, human impacts on behavior, and behavioral shifts due to climate change. Accurate data collection often requires tagging individual animals to collect spatio-temporal state estimates of the animal's geo-position and depth. Acoustic transmitters are prominent due to their continuous communication without requiring retrieval or surfacing to collect data. These transmitters emit underwater acoustic pulses which can be detected by hydrophones. However, the frequent movement of aquatic animals results in high data loss when the animal moves out of the detection range of a stationary hydrophone. Autonomous underwater vehicle (AUV) systems offer a solution for localizing transmitters with higher resolution over longer periods of time. Such systems previously deployed have often required multiple hydrophones mounted on a large frame carried by the AUV. This increases drag, limiting the speed at which the AUV can track highly mobile animals. This work provides an alternative by equipping multiple AUVs with a single compact hydrophone payload. A particle filter algorithm equipped with a hidden Markov model (HMM) behavioral motion model fuses measurements from multiple AUVs to estimate the transmitter's position. Real-world data shows a root mean square error (RMSE) of approximately 10 meters for short-term deployments, and a larger simulated dataset shows an RMSE of approximately 15 meters for longer deployments over a larger area. The HMM fit to historical animal movement data outperforms a generic velocity motion model and both outperform a baseline random walk motion model.
\end{abstract}

\begin{IEEEkeywords}autonomous underwater vehicles, multi-robot systems, source localization, particle filter, hidden Markov model
\end{IEEEkeywords}

\section{Introduction}
\begin{figure*}[t]
    \centering
    \includegraphics[width=4.5in]{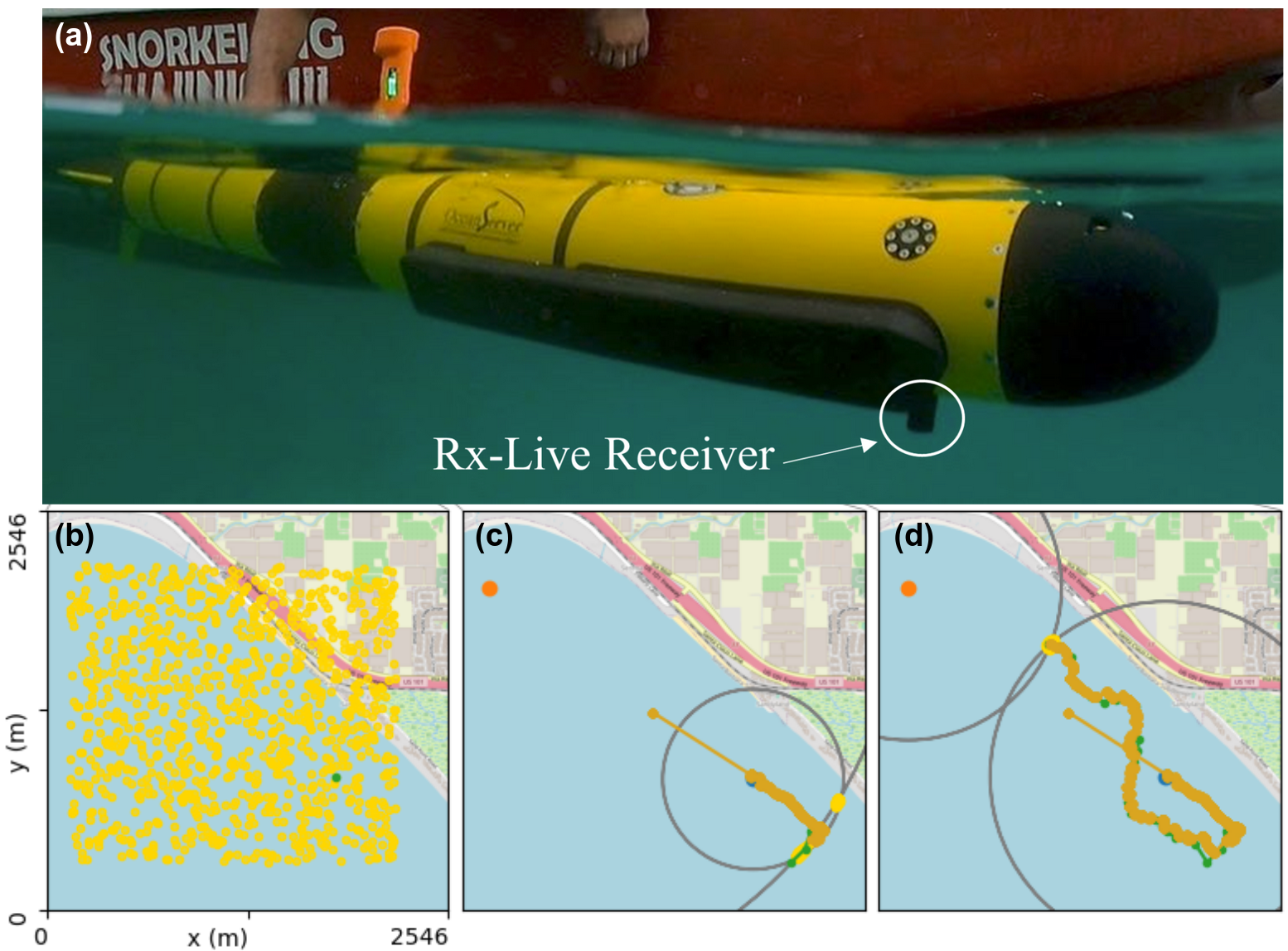}
    \caption{The deployed AUV with Rx-Live Hydrophone Receiver during a field deployment (a) and the particle filter convergence over a trial lasting 50 min (b, c, d). Particles (yellow) converge to an estimated trajectory (gold) within 25 meters of the ground truth trajectory (green) (modified from \cite{Herrera2023}).}
    \label{fig:AUV With Compact Hydrophone}
\end{figure*}

\IEEEPARstart{T}{racking} the movement of aquatic animals has been essential for understanding the behavior, migration patterns, and environmental preferences of a wide range of species. In coastal species, these data have been gathered using acoustic telemetry-based tracking in which acoustic transmitters emit pulse trains encoding measurements. These pulse trains are then received by acoustic receivers, or hydrophones, which detect and decode the signals' data. The emitted signals can be tracked passively by stationary hydrophones with transmitters programmed to emit pulse trains at pseudo-random intervals. They can also be tracked actively by researchers following the animal using a surface-based directional hydrophone system, which requires a constant interval between pulses. Active tracking methods have relied on manual boat based tracking, e.g. \cite{Meese2020,Quinn1991,Shinzaki2013,Metcalfe1997,Holland1992,Lowe2006,Espinoza2011,Filatova2006,Miller2009}, which provides high spatial resolution data of a single individual animal \cite{Lowe2006}, but requires hours of manual labor. On the other hand, passive tracking allows for tracking multiple individuals, but at a lower spatial resolution, and only when the animals are within range of a hydrophone \cite{Espinoza2011}. This range can vary widely depending on the specific transmitter and hydrophone models used, but is typically on the order of a few hundred meters.

Leveraging autonomous underwater vehicles (AUVs) with mounted hydrophones addresses many of the challenges with these approaches. AUVs allow the hydrophones to move with the animal and greatly reduces the manual labor and onsite requirements of the researchers. However, AUVs have limited battery life and have a general increased level of complexity. Additionally, past approaches have utilized AUVs equipped with a large payload holding two hydrophones at each end in order to determine accurate angle of arrival from incoming acoustic signals of tagged animals \cite{Lin2013,Lin2016}. The payload makes state estimation easier but induces a considerable amount of drag leading to reduced battery life and maneuverability. One solution which has been implemented in the past is to enable two way communication between the transmitter and AUV using transponding transmitters \cite{Kukulya2016,Hawkes2020}. This allows for position estimation without the large hydrophone payload, but at the cost of requiring a larger and more complex custom transmitter to be attached to the animal. The transmitter's size limits this technique to large fish, such as basking sharks and white sharks.

In developing a strategy that enables multiple AUVs, each equipped with a single hydrophone, to track marine life affixed with acoustic transmitters, the following contributions have been made:
\begin{enumerate}
    \item An approach to measuring the distance to an off-the-shelf acoustic transmitter using a single omnidirectional hydrophone.
    \item A novel multi-AUV localization algorithm for tracking said transmitter using a hidden Markov model (HMM) as the motion model for a particle filter.
    \item Offline validation of the particle filter with experimental data from real-world ocean deployments, as well as longer trials in simulation.
    \item Comparison of the HMM motion model against a standard velocity motion model and a baseline random walk motion model.
    \item Analysis of the particle filter's convergence with varying measurement loss using in-ocean experimental data, and simulation data.
\end{enumerate}

This paper is organized as follows: section \ref{background} provides relevant background regarding previous work with hydrophones, AUVs, and marine-life tracking methods. Section \ref{system} provides a system overview and describes the multi-robot particle filter. Section \ref{motion_models} introduces the motion models used in the particle filter. Section \ref{datasets} describes the field experiments performed to collect data, as well as the extended trials in simulation. Section \ref{results} presents and analyzes the results of running the particle filter offline and performance with communication loss. Finally, section \ref{conclusions} explains the impact of this work.

\section{Background}
\label{background}

A wide variety of techniques have been used to track marine life with varying success and improvements since the 1990s. Researchers have used satellite transmitters, acoustic telemetry, stationary hydrophone arrays, and manually controlled boats to track the movement of individual animals \cite{Meese2020,Quinn1991,Shinzaki2013,Metcalfe1997,Holland1992,Lowe2006,Espinoza2011,Filatova2006,Miller2009}. However, each of these methods have strong prerequisites for maintaining accurate detection, including ensuring that the animals either remain near the surface of the water (satellite telemetry tracking) or within range of a stationary hydrophone (passive acoustic telemetry tracking). When determining the position of an animal, a manually controlled boat can be used (active tracking), but it requires continuous repositioning of the boat with respect to the movements of the fish below, and the tracking quality is conditional on the human driver’s ability and environmental conditions, e.g. \cite{Shinzaki2013,Metcalfe1997,Holland1992,Lowe2006,Espinoza2011}. In recent years, researchers have developed AUV tracking systems to mitigate the inconsistencies and requirements previously needed to track marine life \cite{Cao2018,Eiler2019,Grothues2008}. For example, AUVs have been equipped with GPS, 3-axis compass, state estimation processors, and a stereo-hydrophone receiver system that listens for acoustic transmitters attached to the marine life being tracked \cite{Shinzaki2013}. The receiver system provides “differential time of arrival data necessary for state estimation,” which is then processed using the particle filter \cite{Forney2012}. These robots were found to have improved spatial accuracy over human-based active tracking as well as a higher frequency of accurate location estimates, with positional errors of less than ten meters \cite{White2015}.

Particle filters, a probabilistic state estimation technique, have gained prominence across various domains for their versatility in handling nonlinear and non-Gaussian systems. In the context of tracking marine life, particle filters have been employed to estimate the state of acoustic transmitters on animals. This methodology can improve the tracking of large mobile marine animals in the wild, where factors like irregular movement patterns, structurally complex habitats, and communication challenges pose significant obstacles. By leveraging particle filters, researchers can effectively fuse data from multiple sources, such as compact omnidirectional acoustic receivers on autonomous underwater vehicles (AUVs), to estimate the position and trajectory of tagged marine animals. 

Probabilistic motion models, in the context of marine life tracking, are models which predict the tagged marine life's change in location over time. These models aid in estimating and predicting trajectories, considering the inherent variability associated with the behavior of marine life. Hidden Markov models (HMMs) are a popular statistical method used to model animal movement, along with other methods such as state-space models and diffusion processes \cite{Patterson2017}. HMMs assume that the animal can be in one of a finite number of behavioral states that cannot be observed directly, and that observable measurements are conditionally independent of past measurements and past states given only the current state. They have been successfully used to model the movement of white sharks \cite{Towner2016}. Simpler random walk motion models are generally considered insufficient to model complex animal movement \cite{Patterson2017}, but offer a useful baseline for comparison to our HMM motion model. Although previous works in the field of marine life tracking have used particle filters \cite{Herrera2023} and hidden Markov models extensively, to the best of the authors' knowledge this is the first work in the field in which both approaches are combined, specifically with the HMM acting as the particle filter's motion model. HMMs and particle filters have been used in conjunction successfully in the context of tracking vehicles using stationary ground sensors and an unmanned aerial vehicle \cite{Ahmed2017}. However, that work is not directly applicable as it can constrain the target vehicles being localized to a much smaller state space \textit{a priori}, i.e. the road network they are driving on.
 
In 2013, Tokekar et. al. \cite{Tokekar2013} studied invasive carp and developed an autonomous robotic boat system to reduce human effort in tracking radio-tagged animals. They used a directional antenna to obtain bearing measurements which were used with an Extended Kalman Filter to estimate transmitter locations.
Clark et. al. \cite{Forney2012} used a particle filter with their AUV in an effort to enable real-time state estimation of sharks and other marine life, including their position, orientation, velocity, and weight, in the immediate area. With multiple AUVs running during testing \cite{Lin2016}, this dramatically increased the accuracy and reduced the mean position error of 25-75\% when determining an animal’s location and movements. However, due to the additional weight and drag of the multi-hydrophone payloads, the AUVs were considerably slower and more difficult to maneuver than without the hydrophones.

To improve the efficiency of the AUVs, the new system replaces the multi-hydrophone payload with a single omnidirectional hydrophone placed within each robot, as shown in figure~\ref{fig:AUV With Compact Hydrophone}. This new hydrophone configuration allows the AUV to move at greater speeds for an increase in mobility and maneuvering, with less cavitation interference. With multiple AUVs in the water and this system in place, larger datasets can be collected in a shorter period of time, thus increasing the efficiency and quantity all the while ensuring the quality of the data. Although not discussed in this paper, prior work on optimal positioning of multiple AUVs for target localization, such as that of \cite{Kehlani2023}, could further improve accuracy. These contributions form a key component for a new system of sensing and an enhanced method of state estimation for highly mobile tagged marine-life animals. This tracking system was validated through real ocean deployments to test the hardware and software which will be discussed throughout the paper.

\section{Multi-Robot System} \label{system}
\subsection{System Overview}

\begin{figure*}[t]
  \centering
  \includegraphics[width=4.5in]{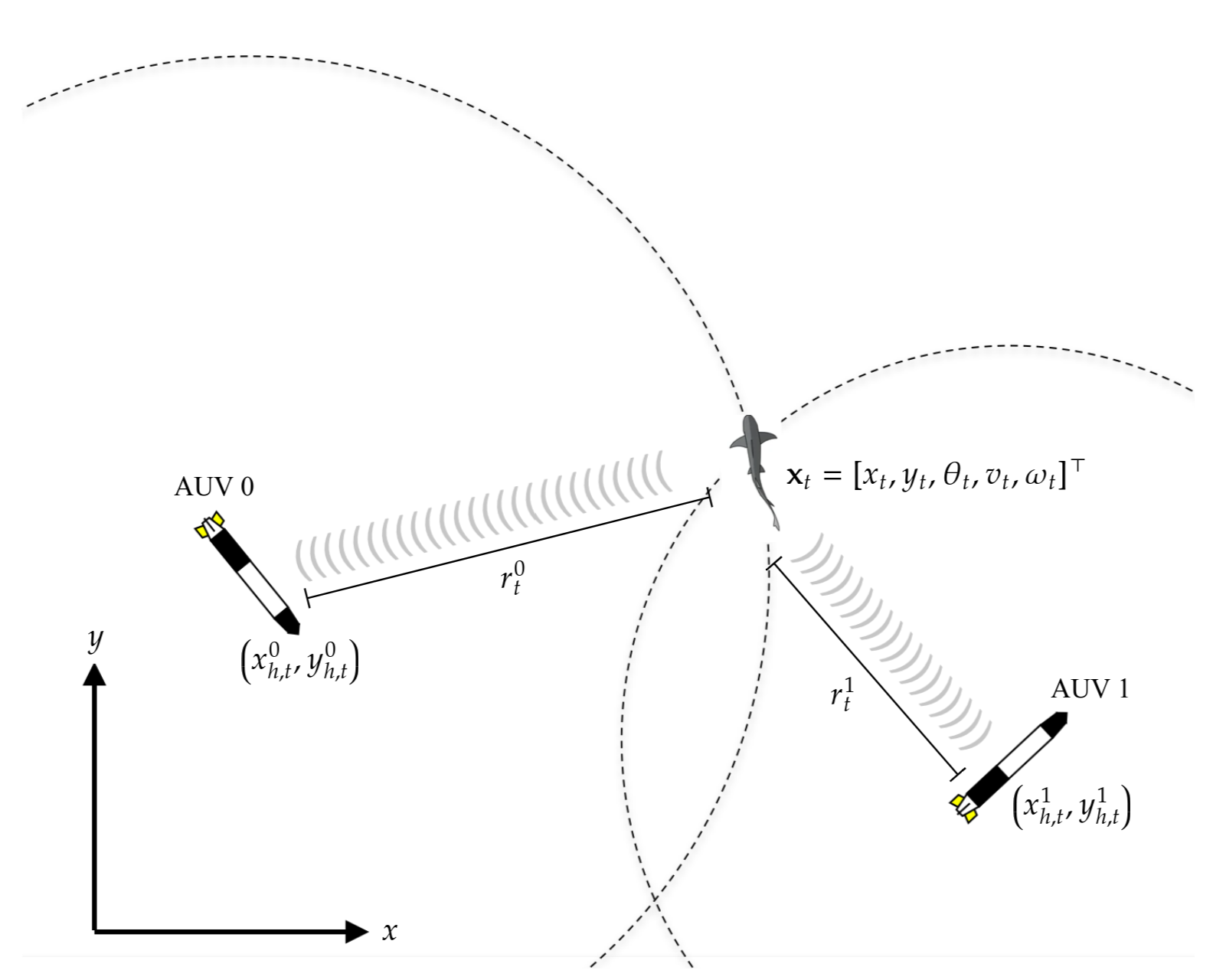}
  \caption{Top down view of two Rx hydrophone receiver equipped AUVs estimating the geoposition of an acoustically tagged shark. The AUV's coordinates at time $t$ are denoted as $x^0_{h,t}$, $y^0_{h,t}$ for AUV 0 and $x^1_{h,t}$, $y^1_{h,t}$ for AUV 1. The transmitter's state vector at time $t$, $\textbf{x}_t$, consists of the transmitter's coordinates $x_t$, $y_t$, its heading $\theta_t$, and its linear and angular velocities $v_t$, $\omega_t$ (modified from \cite{Herrera2023}).}
  \label{fig:diagram}
\end{figure*}

The multi-robot system consists of multiple AUVs and an acoustic transmitter. Figure~\ref{fig:diagram} shows a top down view with the AUVs positioned around the transmitter, which is attached to the shark. The animal borne transmitter emits a pulse train (from here on called a signal) that encodes the transmitter's ID, and for some types of transmitters, various sensor data (e.g. temperature, depth, summed acceleration). The AUVs are capable of detecting and decoding the transmitter's signals via mounted omnidirectional hydrophones.

While the system is 3 dimensional, for the current work it is assumed that the AUVs and the transmitter remain close to the surface at all times, allowing the use of 2D positions and orientations. During field experiments the transmitter and hydrophones were deployed in shallow water and stayed close to the surface throughout, validating this assumption. This assumption was also validated using a large dataset of white shark trajectories described in section \ref{simulated_dataset}, in which the sharks stayed within 10.7 meters of the water's surface.
For the general case where full 3D state estimation is needed, acoustic transmitters that include depth sensors have successfully been used for 3D position state estimation in other AUV tracking work, e.g. \cite{Lin2014,Lin2016}.
The depth measurements allow the distance along the surface of the water to the shark's location to be calculated, which effectively reduces the state estimation problem to the 2D case. Simulated data with depth measurements were used to demonstrate this, which is discussed further in section \ref{depth_correction}.

The AUV's locations and velocities are assumed to be known via on board sensors (GPS, compass, DVL), but the transmitter's state is unknown. At time $t$, the state $\textbf{x}_t$ to be estimated consists of the transmitter's coordinates $x_t$, $y_t$, its orientation $\theta_t$, its linear speed $v_t$, and its angular velocity $\omega_t$
\begin{align}
    \textbf{x}_t &= [x_t, y_t, \theta_t, v_t, \omega_t]^\top.
\end{align}

\subsection{Distance Estimation} \label{distance_estimation}

The first step to estimating the state of the transmitter (i.e. the tagged shark) is estimating the distances between it and each AUV. The type of transmitter used in this work emits signals at a constant rate of approximately $0.125$ Hz, or once every $8$ seconds, with the exact value depending on the specific transmitter. The hydrophone mounted on the AUV actively listens for these signals and has some probability of detecting them depending on the distance, the amount of ambient noise, and whether it has a clear path to the transmitter. When a detection occurs, the hydrophone receiver records the transmitter's ID, an estimate of the signal's strength, and the time at which the detection occurred. If the signal's time of flight (TOF) were known, then the distance $r$ between the hydrophone and transmitter could be computed by multiplying the TOF by the speed of sound in water ($s=1460$ m/s). Unfortunately the TOF cannot be measured directly, but the amount by which the TOF has changed between detections, $\Delta\text{TOF}$, can be used.

Let $t_i$ denote the time at which the $i$th detection was recorded. By comparing the difference in time between consecutive detections, $\Delta t_{i-1:i}=t_i-t_{i-1}$, with the transmitter's known period $T$ between signals, it is possible to compute $\Delta\text{TOF}_{t_{i-1}:t_i}$, i.e. the change in TOF between the $i$th detection at time $t_i$ and the previous detection at time $t_{i-1}$. If $\Delta t_{i-1:i}$ is greater than $T$, this indicates that the transmitter has moved further from the hydrophone in the time between the two detections, and vice versa:
\begin{align}
    \Delta\text{TOF}_{t_{i-1}:t_i} = \Delta t_{i-1:i}-T.
\end{align}
This equation assumes that all signals are detected. If a signal is not detected, then $\Delta t_{i-1:i}$ would be close to $2T$, implying that the transmitter moved away a very large distance, on the order of $sT\approx10000$ m, which is unrealistic. Therefore we instead make the assumption that if $\Delta t_{i-1:i}>\frac{1}{2}T$, then a signal was missed. Modding out by $T$ accounts for an arbitrary number of missed signals between detections $i-1$ and $i$:
\begin{align}
    \Delta\text{TOF}_{t_{i-1}:t_i} = \text{mod}\left(\Delta t_{i-1:i}-\frac{1}{2}T,T\right)-\frac{1}{2}T.
\end{align}
Changes in the hydrophone's distance from the transmitter between detections $\Delta r_t$ can then be computed as:
\begin{align}
    \Delta r_{t_{i-1}:t_i} &= r_{t_i} - r_{t_{i-1}} = s\Delta\text{TOF}_{t_{i-1}:t_i}, \\
    \Delta r_{t_0:t_i} &= r_{t_i}-r_{t_0} = \Delta r_{t_0:t_1}+\cdots+\Delta r_{t_{i-1}:t_i}.
\end{align}
Taking $i$ to refer to the latest detection as of time $t$, the total change in distance between the transmitter and the hydrophone as of time $t$ is $\Delta r_{0:t}=\Delta r_{0:i}$.

The subsequent results presented assume that the initial distance $r_0$ between the transmitter and the AUV is known. This can be accomplished by starting the system with the transmitter next to the hydrophone. In that case, $r_0=0$, which means $\Delta r_{0:t}=r_t$. This allows the absolute distance between the transmitter and the AUV to be estimated for the duration of the experiment. However, accumulating the changes in distance in this way introduces small errors that cause the distance estimate to drift over time. In short term experiments, such as those with similar time frames to the validation experiments discussed in this paper, the drift in the accumulated distance estimate can be ignored. In longer term experiments, if the transmitter can be recovered, the distance estimates can be re-calibrated offline using a known final distance. Examples of previous work where a recoverable transmitter was used include \cite{Lin2016}. Future work is needed for long-term experiments in which the transmitter cannot be recovered, or in experiments where a known initial distance cannot be obtained. For example, this work could be extended to incorporate knowledge of the distance-dependent probability of detection as in \cite{Guo2026}.

\subsection{Particle Filter}
\begin{algorithm}[h]
\caption{Particle Filter Initialization($n$)}
\label{alg:pf_init}
\begin{algorithmic}[1]
    \STATE // Initialize particles uniformly randomly
    \FOR{$i=1$ to $n$}
        \STATE sample $x^1_0,y^i_0 \sim U(x_\text{min},x_\text{max}),U(y_\text{min},y_\text{max})$ 
        \STATE sample $\theta^i_0 \sim U(-\frac{\pi}{2},\frac{\pi}{2})$
        \STATE $p^i_0=\begin{bmatrix}x^i_0 & y^i_0 & \theta^i_0 & 0 & 0 & 1\end{bmatrix}^\top$
    \ENDFOR
    \STATE $P_0=\{p^1_0,\dots,p^n_0\}$
\end{algorithmic}
\end{algorithm}
The particle filter is a state estimation algorithm that uses a collection of particles to represent a probability distribution over the state. It consists of a prediction and correction step. While no new measurements are available, the prediction step propagates particles forward in time according to a motion model. As measurements are received, the correction step assigns weights to each particle using a measurement model and resamples the particles. Resampling consists of drawing a new set of particles from the old set with probabilities proportional to the weights and with replacement. One important feature of particle filters is their ability to represent multi-modal state distributions by using a non-parametric representation. This feature is desirable for this system because there are cases where two distinct locations are plausible before convergence. It can also represent a uniform distribution at the start if an initial state is not known by initializing the particles uniformly randomly. Algorithm~\ref{alg:pf_init} provides pseudocode for initializing the particle filter, and algorithm~\ref{alg:pf_it} provides pseudocode for a particle filter step.

\begin{algorithm}[t]
\caption{Particle Filter Iteration($P_t$, $Z_t$)}
\label{alg:pf_it}
\begin{algorithmic}[1]
    \STATE // Prediction step
    \FOR{$p^i_t$ in $P_t$}
        \STATE sample $p^i_{t+1} \sim Pr(\textbf{x}_t|p^i_t)$  // Implemented by motion model
    \ENDFOR
    \STATE $P_{t+1} = \{p^1_{t+1},\dots,p^n_{t+1}\}$
    \STATE
    \STATE // Correction step
    \IF{$Z_t\neq\emptyset$}
        \FOR{$p^i_t$ in $P_t$}
            \STATE $w^i_t = Pr(Z_t|p^i_t)$  // equation~(\ref{eq:reweight})
        \ENDFOR
        \STATE
        \STATE // Resample
        \STATE $P_{t+1}=\emptyset$
        \FOR{$i=1$ to $n$}
            \STATE draw $i$ from $\{1,\dots,n\}$ with probability $\propto w^i_t$
            \STATE add a copy of $p^i_t$ to $P_{t+1}$ with weight set to $1$
        \ENDFOR
    \ENDIF
\end{algorithmic}
\end{algorithm}
Particle $i$ at time $t$ and the set of particles at time $t$ are denoted respectively as
\begin{align}
    p^i_t &= \begin{bmatrix}x^i_t & y^i_t & \theta^i_t & v^i_t & \omega^i_t & w^i_t\end{bmatrix}^\top, \\
    P_t &= \{p_t^1,\dots,p_t^n\}.
\end{align}
where $w$ is the particle's weight and $n$ is the number of particles. A measurement $z_t$ at time $t$ consists of the distance $r$ between a hydrophone and the transmitter, and its time derivative $\dot{r}$
\begin{align}
    z_t=\begin{bmatrix}r_t & \dot{r}_t\end{bmatrix}^\top,
\end{align}
and its covariance matrix is denoted
\begin{align}
    \Sigma_z=\begin{bmatrix}
        \text{var}(r_t) & \text{cov}(r_t,\dot{r}_t) \\
        \text{cov}(\dot{r}_t,r_t) & \text{var}(\dot{r}_t)
    \end{bmatrix}.
\end{align}
Measurements are obtained for each detection using the TOF method described in section~\ref{distance_estimation}, and the covariance matrix can be calculated from sample data collected separately.

\subsubsection{Prediction Step}
Algorithm 2, lines~2-5 are the particle filter's prediction step. For each particle (line 2), the motion model is used to propagate the particles forward in time, updating their positions, orientations, and velocity (line 3). The particle's weights are not updated until the correction step. The $i$th particle at time $t+1$ is given by
\begin{align}
    p^i_{t+1} &= \begin{bmatrix}x_{t+1} & y_{t+1} & \theta_{t+1} & v_{t+1} & \omega_{t+1} & w_t\end{bmatrix}^\top.
\end{align}
See section \ref{motion_models} for details on how each of the three motion models tested updates the particles. Prediction steps are scheduled at regular $1$ second intervals.

\subsubsection{Correction Step}

Lines 7-19 are the particle filter's correction step. The set of $m$ measurements available at the current time step is denoted as $Z_t=\{z^0_t,\dots,z^m_t\}$. If both hydrophones detected the same signal from the transmitter, two measurements will be available at the current time step. Otherwise, only one or zero measurements may be available. If $Z_t$ is not empty (line~8), then the particles are weighted by comparing the true measurements $z^j_t$ with the predicted measurements $\hat{z}^{i,j}_t$ given the $i$th particle's state (line~10). The predicted distance and relative speed for particle $i$ and measurement $j$ are given by
\begin{align}
    \hat{r}^{i,j}_t &= \sqrt{(x^j_{h,t}-x^i_t)^2+(y^j_{h,t}-y^i_t)^2}, \\
    \hat{\dot{r}}^{i,j}_t &= \frac{(x^j_{h,t}-x^i_t)(\dot{x}^j_{h,t}-\dot{x}^i_t)+(y^j_{h,t}-y^i_t)(\dot{y}^j_{h,t}-\dot{y}^i_t)}{\hat{r}^{i,j}_t}.
\end{align}
where $x^j_{h,t}$, $y^j_{h,t}$, $\dot{x}^j_{h,t}$, and $\dot{y}^j_{h,t}$ are the $x$ and $y$ coordinates and $x$ and $y$ velocities respectively of the hydrophone $h$ that produced measurement $j$ at time $t$. The predicted measurements are then used to update the weights as follows
\begin{align}
    w^i_t = \Pi^m_{j=1}\mathcal{N}(\hat{z}^{i,j}_t|z^j_t,\Sigma_z). \label{eq:reweight}
\end{align}
where $\mathcal{N}(x|\mu,\Sigma)$ denotes the Gaussian distribution with mean $\mu$ and covariance $\Sigma$.
The final weight is the joint probability of all of the measurements, assuming independence.

After re-weighting, the particles are resampled. New particles are randomly drawn with replacement from the current set of particles (line 16), which becomes the new set of particles in the next iteration (line 17).

\section{Motion Models} \label{motion_models}
Motion models are used in the prediction step of the particle filter to propagate the particles' states forward in time. Three different motion models were tested: the baseline motion model, the velocity motion model, and the HMM motion model.

\subsection{Baseline Motion Model}

To update the particle state, random numbers $\Delta x$ and $\Delta y$ are sampled from Gaussians with mean $0$ and are added to each particle's current $x_t$ and $y_t$ coordinates to produce $x_{t+1}$ and $y_{t+1}$. The particles' new angles and new velocities are calculated using the old and new coordinates:

\begin{align}
    \theta_{t+1} &= \text{atan2}(\Delta y, \Delta x), \\
    v_{t+1} &= \frac{\sqrt{\Delta x^2 + \Delta y^2}}{\Delta t}, \\
    \omega_{t+1} &= \frac{\theta_{t+1}-\theta_t}{\Delta t}.
\end{align}

The new particles' weights are set to the old particles' weights.

\subsection{Velocity Motion Model}

In the velocity motion model, random numbers $\Delta v$ and $\Delta\omega$ are sampled from Gaussians with $0$ mean and are added to each particle's current linear velocity $v_t$ and angular velocity $\omega_t$ to produce $v_{t+1}$ and $\omega_{t+1}$.

The particles' new coordinates and orientation are then calculated as follows:
\begin{align}
    x_c &= x_t - \frac{v_{t+1}}{\omega_{t+1}}\sin(\theta_t) \label{eq:mm_xc}, \\
    y_c &= y_t + \frac{v_{t+1}}{\omega_{t+1}}\cos(\theta_t) \label{eq:mm_yc}, \\
    x_{t+1} &= x_c + \frac{v_{t+1}}{\omega_{t+1}}\sin(\theta_t + \omega_{t+1} dt), \label{eq:mm_x}\\
    y_{t+1} &= y_c - \frac{v_{t+1}}{\omega_{t+1}}\cos(\theta_t + \omega_{t+1} dt), \label{eq:mm_y}\\
    \theta_{t+1} &= \theta_t + \omega_{t+1} dt \label{eq:mm_theta}.
\end{align}
The new particles' weights are set to the old particles' weights.

\subsection{HMM Motion Model}
\begin{figure*}
    \centering
    \includegraphics[height=2.5in]{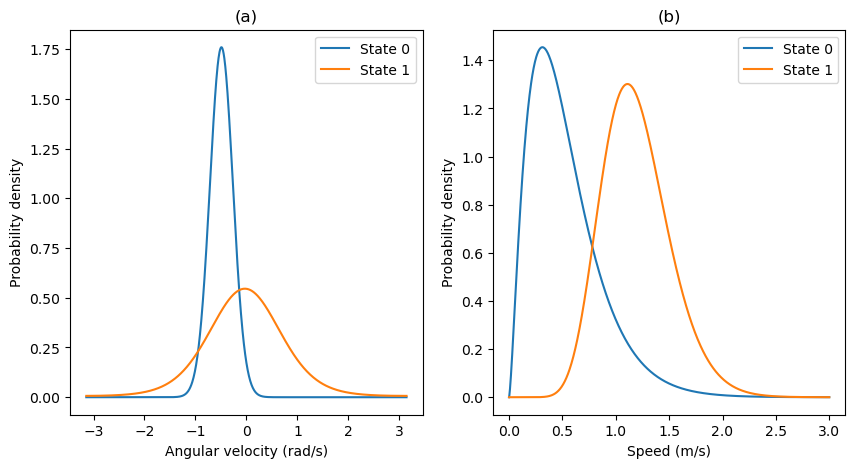}
    \caption{State dependent probability distributions for the angular velocity (a) and linear velocity (b).}
    \label{fig:emissions}
\end{figure*}
A two state hidden Markov model (HMM) was fitted to the white shark trajectory data described in section \ref{simulated_dataset}. The linear and angular velocities of the sharks were used as the observable variables, and were assumed to be distributed according to gamma and von Mises distributions respectively. The observables' state dependent distributions are plotted in figure \ref{fig:emissions}. In addition to the state variables used by the other two motion models, the HMM motion model also keeps track of the HMM state for each particle, which is updated by sampling from the HMM's state transition probabilities. The HMM state is used to sample from the state dependent distributions which become the particles' next linear and angular velocities. After sampling the new linear and angular velocities, the HMM motion model uses the same equations (\ref{eq:mm_xc}-\ref{eq:mm_theta}) as the velocity motion model to update the particles' coordinates and orientation.

\section{Datasets} \label{datasets}

\begin{figure*}[t]
    \centering
    \begin{subfigure}[b]{0.49\textwidth}
        \includegraphics[height=2.5in]{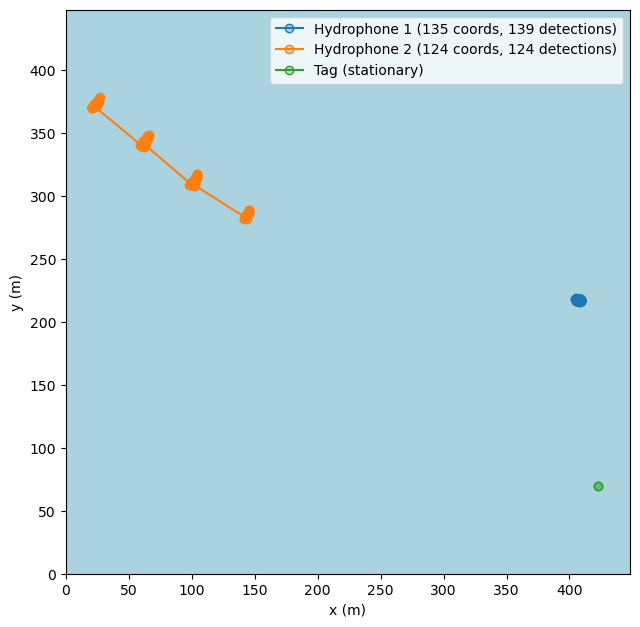}
        \caption{}
    \end{subfigure}
    \begin{subfigure}[b]{0.49\textwidth}
        \includegraphics[height=2.5in]{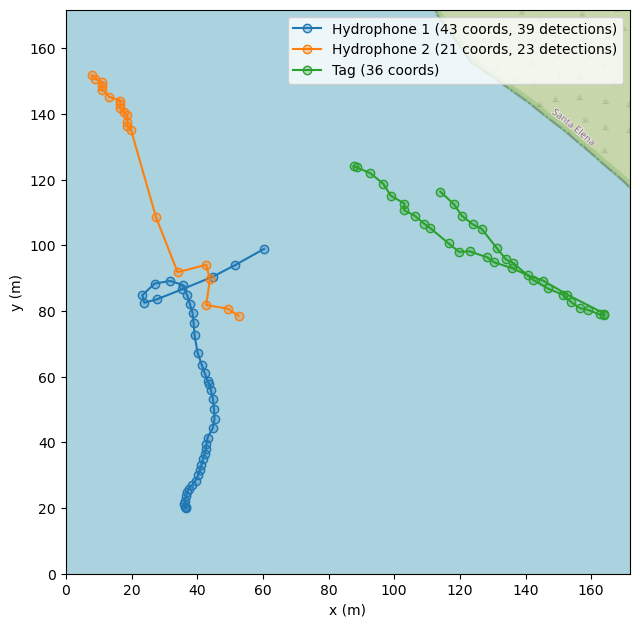}
    \caption{(modified from \cite{Herrera2023})}
    \end{subfigure}
    \begin{subfigure}[b]{0.49\textwidth}
        \includegraphics[height=2.5in]{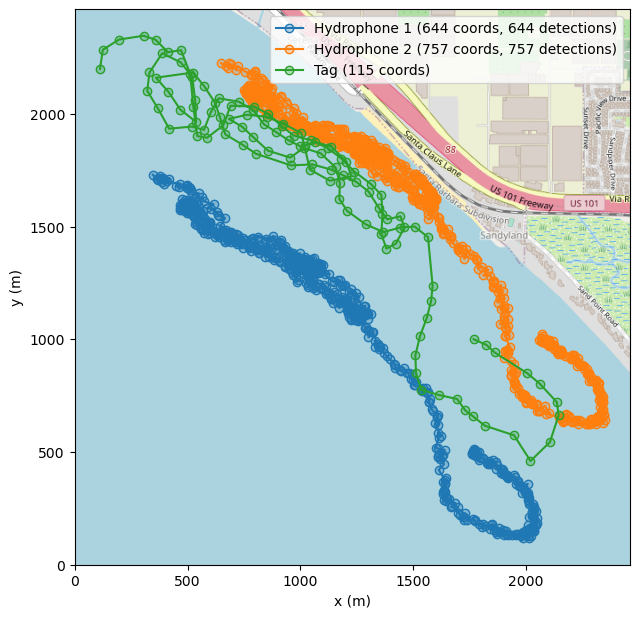}
    \caption{}
    \end{subfigure}
    \caption{Trajectories of the hydrophones and transmitter in the Long Beach (a), Santa Elena Bay (b), and simulated (c) datasets.}%
    \label{fig:dataset_overviews}%
\end{figure*}

To collect data to validate the particle filter, the hydrophones and acoustic transmitter were deployed off the coast of Long Beach, CA and in Santa Elena Bay on the north Pacific coast of Costa Rica.

\subsection{Long Beach Dataset}
The first dataset was collected on 8 June, 2022 in Long Beach, CA. During the collection of this dataset, one of the hydrophones was deployed from a boat that was anchored in place, while the other was mounted on an AUV. The transmitter used had a constant pulse interval of 8.179 s and was also anchored in place using a buoy. While the boat and transmitter were stationary, the AUV drove progressively further from the transmitter in 50 meter increments. Transmitter detections and coordinates of the hydrophones were recorded for 1200 seconds. Since the transmitter was stationary, its coordinates were only recorded once when the buoy was deployed. The trajectory of the AUV mounted hydrophone and the locations of the other stationary hydrophone and transmitter can be seen in Figure~\ref{fig:dataset_overviews} (top left) in orange, blue, and green respectively.

\subsection{Santa Elena Bay Dataset}
The second dataset was collected on 21 July, 2022 in Santa Elena Bay, Costa Rica. During the collection of this dataset, both hydrophones were deployed from boats while the transmitter was moved near the shore of Santa Elena bay by a swimmer. The transmitter used was the same one as the Long Beach dataset, with a constant pulse interval of 8.179 s. The boats were allowed to drift for most of the deployment but were moved when necessary to stay near the transmitter. Both hydrophones as well as the transmitter moved continuously throughout this deployment, which lasted about an hour. As with the Long Beach dataset, the hydrophones' continuously recorded transmitter detections and their own coordinates. The transmitter's coordinates were recorded continuously by a waterproof GPS carried by the swimmer. There were substantial gaps in the transmitter's coordinates resulting in long periods without groundtruth data. For that reason, the Santa Elena Bay dataset consists of only 349 seconds, which was the longest span of time during which the transmitter's coordinates were recorded and there were a substantial number of detections by both hydrophones. The trajectories of the two hydrophones (orange and blue) and transmitter (green) can be seen in Figure~\ref{fig:dataset_overviews} (top right).

\begin{figure*}[t]
    \centering
    \begin{subfigure}[b]{0.49\textwidth}
        \includegraphics[height=2.2in]{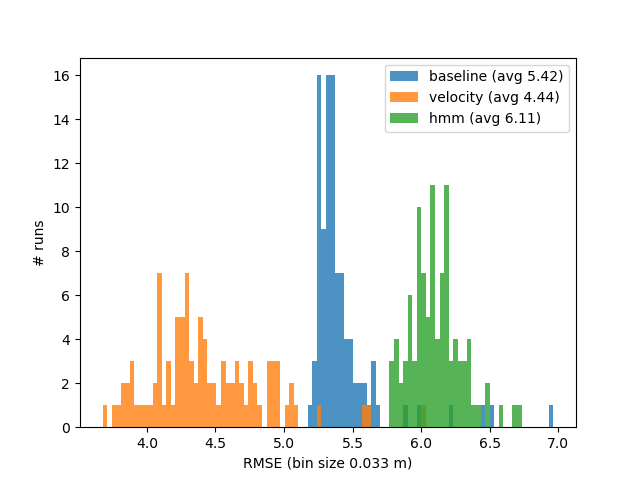}%
        \caption{}
    \end{subfigure}
    \begin{subfigure}[b]{0.49\textwidth}
        \includegraphics[height=2.2in]{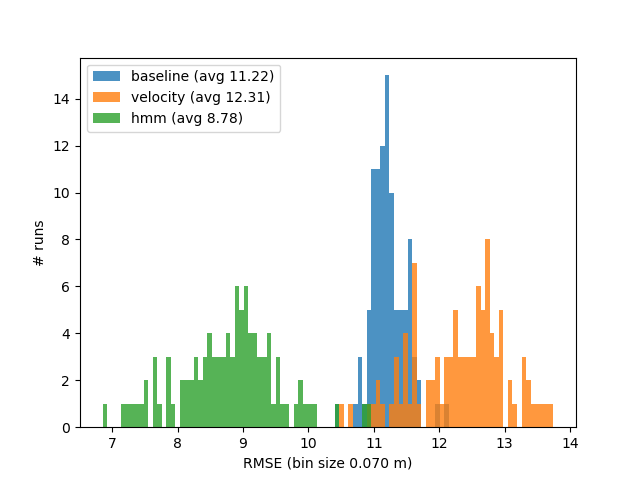}%
        \caption{}
    \end{subfigure}
    \begin{subfigure}[b]{0.49\textwidth}
        \includegraphics[height=2.2in]{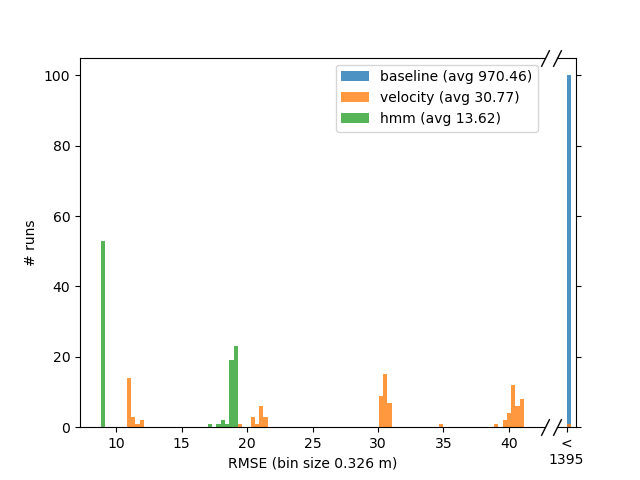}%
        \caption{}
    \end{subfigure}
    \caption{Results of 100 particle filter trials for each motion model for the Long Beach (a), Santa Elena Bay (b), and simulated (c) datasets.}%
    \label{fig:mm_comparisons}
\end{figure*}

\subsection{Simulated Dataset} \label{simulated_dataset}
To augment the real data collected from the deployments in Long Beach and Santa Elena Bay, a simulated dataset was created in order to test the particle filter on longer trials. This simulated dataset was based off of a large dataset (around 124k position observations) of white shark trajectories collected by the Shark Lab at California State University, Long Beach. The Shark Lab placed twenty-four static hydrophone receivers along the coast of Santa Barbara, CA to track the positions of tagged sharks. From the months of May 2020 to August 2020, twenty two white sharks were tracked. Given the time of detection from received transmitter signals, the proprietary VEMCO Positioning System was used to derive the shark's positions. Due to dropped transmitter signals, which is a common phenomenon when passively tracking tagged animals in the wild, this resulted in a large collection of small trajectory segments for each shark. These data were filtered for trajectory segments in which consecutive detections occurred no more than 3 min apart. The longest such trajectory segment consists of 115 white shark coordinates and lasts 150 min. The shark's coordinates in this dataset were linearly interpolated to match the particle filter's 1 second frequency. The two hydrophones' simulated paths were obtained by shifting the shark's trajectory by a constant offset and then smoothing it using a low-pass filter. The effect is that the hydrophones' paths lag behind the shark's, which simulates them maneuvering towards it throughout the trial. While their distance to the shark varies as their path lags behind it, they are arranged in such a way as to keep the angle between them and the shark at roughly 90 degrees (shown in figure~\ref{fig:sim_dataset_pf_run_error}), which helps minimize the error in the estimated location. Gaussian noise was added to the hydrophone's simulated coordinates based off of the real GPS's resolution. The simulated transmitter emitted a signal every 8 seconds, with Gaussian noise based off of the real transmitter's period's noise. The transmitter's signals were simulated as having a fixed probability of being dropped per meter distance that they traveled, resulting in an exponential drop off probability as distance increases. This resulted in 645 out of the 1125 simulated signals being dropped for one hydrophone, and 671 out of 1125 for the other. This simulated dataset allowed the particle filter to be tested with a much longer trial than the data collected from the deployments. The actual trajectory taken by the shark (green), as well as the simulated hydrophone locations (orange and blue) can be seen in Figure~\ref{fig:dataset_overviews} (bottom).

\section{Results} \label{results}

The particle filter was run on all 3 datasets with all 3 motion models. Sections \ref{long_beach_results}, \ref{santa_elena_bay_results}, and \ref{sim_results} present the results of 300 particle filter trials, with 100 trials per motion model. For each trial, the position estimate error is plotted over time. The position estimate error at a given time $t$ is defined as the distance between the transmitter's estimated position $\hat{x}_t,\hat{y}_t$ and groundtruth position $x_t,y_t$:
\begin{align}
    \text{pos est error}_t &= \sqrt{(x_t-\hat{x}_t)^2+(y_t-\hat{y}_t)^2}.
\end{align}
To account for high position estimate errors at the start of the trial before the particles converge, the RMSE is taken using the position estimate errors after convergence:
\begin{align}
    \text{RMSE} &= \frac{1}{T-t_c}\sum^T_{t=t_c}\sqrt{(x_t-\hat{x}_t)^2+(y_t-\hat{y}_t)^2}.
\end{align}
where $t_c$ is the time to convergence. In order to capture the intuitive notion of convergence as the particles being close together spatially, the time to convergence is defined in terms of the particles' $x$ and $y$ covariance matrix:
\begin{align}
    \Sigma_{xy} &= \begin{bmatrix}\text{var}(x) & \text{cov}(x,y) \\ \text{cov}(y,x) & \text{var}(y)\end{bmatrix}.
\end{align}
The covariance matrix's eigenvector with the largest eigenvalue gives the direction with the largest variance, and the largest eigenvalue itself is that variance:
\begin{equation}
    \begin{split}
        \lambda_1 &= \frac{\text{var}(x)+\text{var}(y)}{2} \\
        &+ \sqrt{\left(\frac{\text{var}(x)-\text{var}(y)}{2}\right)^2+\text{cov}(x,y)^2}.
    \end{split}
\end{equation}
The time to convergence is then defined as the first time when $\sqrt{\lambda_1}$ (i.e. the standard deviation in the direction with the largest variance) falls below a threshold, which was set to $10$ m for the following results, or $0$ if the filter never converged. For each motion model, the RMSE after convergence is calculated for each of the 100 trials, and the average RMSEs are reported. For each dataset, specific trials are also presented in order to highlight the behavior of the particle filter and motion model.

\begin{figure*}[t]
    \centering
    \includegraphics[width=4.5in]{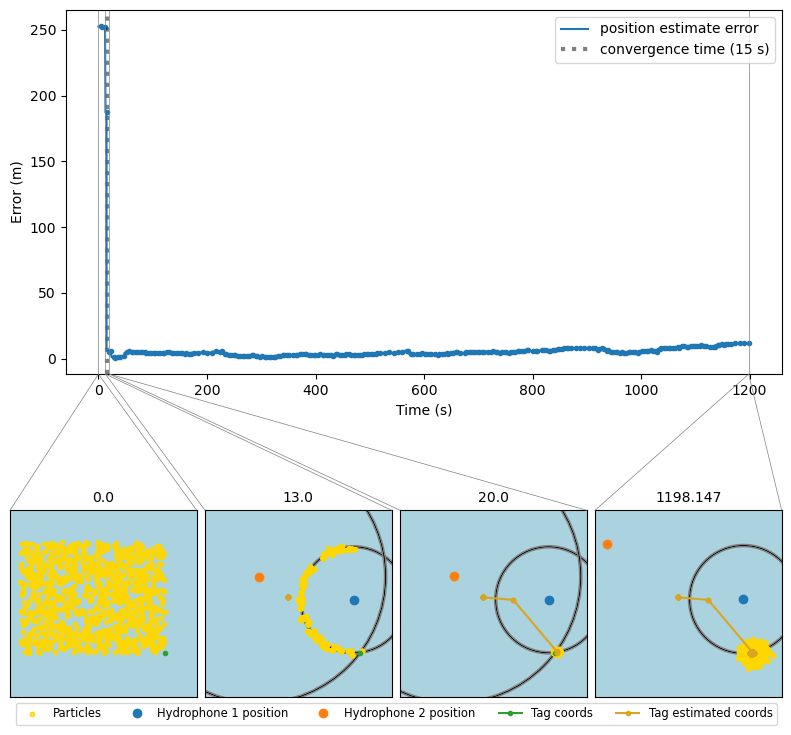}%
    \caption{Error for a single particle filter trial using the HMM motion model on the Long Beach dataset, and particle distributions at different times during the trial.}
    \label{fig:long_beach_pf_run_error}
\end{figure*}

\subsection{Long Beach Dataset Results} \label{long_beach_results}

The Long Beach dataset had the simplest movement of the 3 datasets used, with only a single hydrophone moving in 50 meter intervals and a stationary transmitter. All 3 models performed very similarly, as can be seen in figure \ref{fig:mm_comparisons} (top). The average RMSEs for the baseline, velocity, and HMM motion models were 5.42 m, 4.44 m, and 6.11 m respectively.

Figure \ref{fig:long_beach_pf_run_error} plots the error and highlights the state of the particle filter at 4 different times for a single trial of the particle filter using the HMM motion model. The 4 bottom subfigures show the particles' distributions in yellow and the transmitter's estimated trajectory in gold. The hydrophones are shown in blue and orange, and the hydrophone measurements (i.e. the measured distances to the transmitter) are shown as circles around the hydrophones. At 0 s, the particles are initialized randomly. After 13 s the particles have concentrated in a partial circle around one of the hydrophones. After only 20 s, the particles have converged at the transmitter's true location. Because of the simple movement of the hydrophones and transmitter, all 3 motion models are capable of keeping particles around the transmitter's location. The particles remain converged around the transmitter's location throughout the rest of the dataset, with only a slight increase in error as the particles spread out, as can be seen at the end of the data after 20 min (1,198 s).

\subsection{Santa Elena Bay Dataset Results} \label{santa_elena_bay_results}
\begin{figure*}[t]
    \centering
    \includegraphics[width=4.5in]{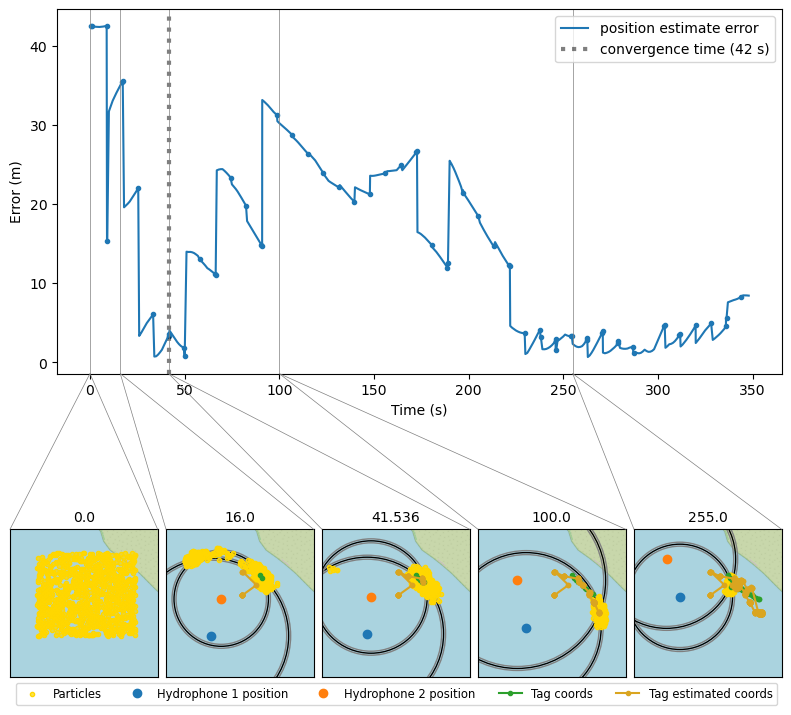}%
    \caption{Error for a single particle filter trial using the HMM motion model on the Santa Elena Bay dataset, and particle distributions at different times during the trial (modified from \cite{Herrera2023}).}
    \label{fig:santa_elena_bay_pf_run_error}
\end{figure*}

The Santa Elena Bay dataset had more complex movement of both the transmitter and the hydrophones. As can be seen in figure \ref{fig:mm_comparisons} (middle) the baseline and velocity motion models performed similarly, while the HMM motion model did slightly better. The average RMSEs for the baseline, velocity, and HMM motion models were 11.22 m, 12.31 m, and 8.78 m respectively.

Figure \ref{fig:santa_elena_bay_pf_run_error} plots the error and highlights the state of the particle filter at 5 different times for a single trial of the particle filter using the HMM motion model using the same format as figure \ref{fig:long_beach_pf_run_error}. Once again, the particles are initialized randomly at 0 s. After 16 s, the particles are concentrated around the overlap between the two hydrophone measurements. After 41 s, the particles have for the most part converged around the true transmitter location, although some particles remain around the other intersection point of the two hydrophone measurements. After 100 s the particles have fully converged around the transmitter's true location after more measurements from both datasets, although there is still some error. Finally, after 255 s, the particles have converged much closer to the transmitter's true location.

It is worth noting that in many of the trials, the HMM motion model takes much longer to converge than either the baseline or velocity motion models. The average convergence times are 27~s, 40~s, and 180~s for the baseline, velocity, and hmm motion models respectively. In trials where the HMM motion model takes longer to converge, the cause is that particles remain around both intersection points of the two measurements throughout more of the trial than in the example trial shown in figure \ref{fig:santa_elena_bay_pf_run_error}.

\subsection{Simulated Dataset Results} \label{sim_results}
\begin{figure*}[t]
    \centering
    \includegraphics[width=4.5in]{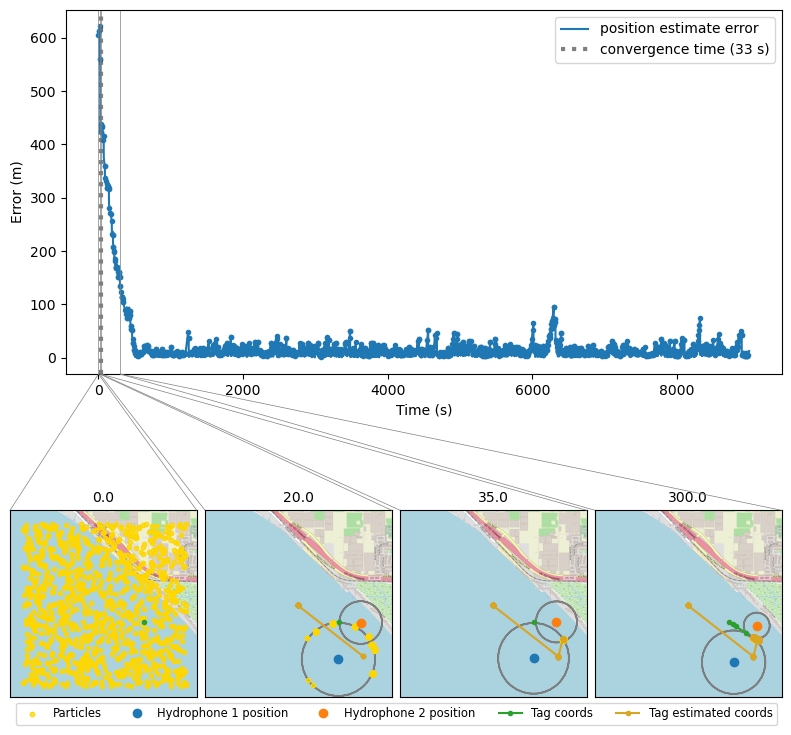}%
    \caption{Error for a single particle filter trial using the velocity motion model on the simulated dataset, and particle distributions at different times during the trial.}
    \label{fig:sim_dataset_pf_run_error}
\end{figure*}

The simulated dataset was the largest both in terms of the duration and the distances covered by the transmitter's movement. As can be seen in figure \ref{fig:mm_comparisons} (bottom) the HMM motion model performed the best, followed by the velocity motion model, followed by the baseline motion models which performed significantly worse than the other two. The average RMSEs for the baseline, velocity, and HMM motion models were 970.46 m, 30.77 m, and 13.62 m respectively.

Figure \ref{fig:sim_dataset_pf_run_error} plots the error and highlights the state of the particle filter at four different times for a single trial of the particle filter, using the same format as figures \ref{fig:long_beach_pf_run_error} and \ref{fig:santa_elena_bay_pf_run_error}. Once again the particles are initialized randomly at 0 s. 20 s into the trial, the particles have concentrated in a partial circle around one of the hydrophones. At 35 s into the trial, due to the angle between the transmitter and hydrophones, the measurements yield two intersection points. As a result, the particles have converge away from the transmitter's true location, and the RMSE remains high even after convergence. This accounts for the two peaks seen in the HMM's RMSE histogram: when the particles converge around the true location, the RMSE is low, and when they converge around the wrong location, the RMSE is high. 300 s into the trial, the transmitter and hydrophones form a shallower angle, so the hydrophone measurements' two intersection points are relatively close to one another and the filter can converge around the true location. This occurs occasionally throughout the trial, and when the angle once again approaches 90 degrees, the two intersection points diverge. The HMM motion model is able to maintain particle diversity at both intersection points until enough data are gathered to distinguish between the two, although this can cause the error to spike as the average particle location lies between the two intersection points. Although the velocity motion model can also do this, it does worse than the HMM motion model at distinguishing the two intersection points, as shown by the four distinct peaks in its RMSE histogram. The baseline motion model cannot maintain particles around the true location for very long and diverges almost all of the time, explaining its much larger RMSEs. Although not shown in this simulated trial, with different simulated hydrophone trajectories, the velocity motion model can diverge in the same way as the baseline motion model, leading to much larger RMSEs than the HMM motion model.

\subsection{Measurement Loss} \label{measurement_loss}

\begin{figure*}[t]
    \centering
    \includegraphics[width=4.5in]{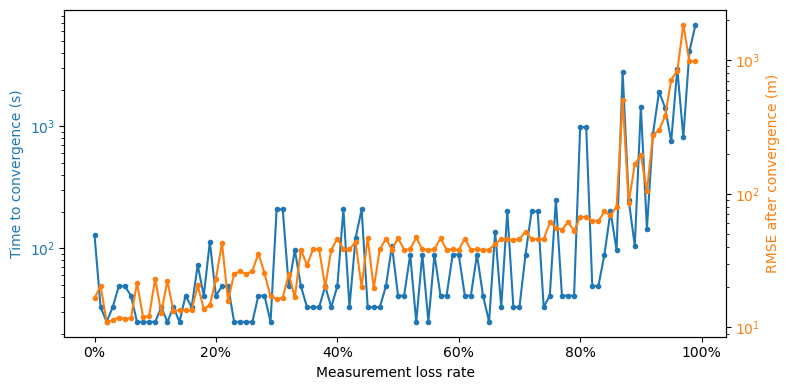}%
    \caption{Time to convergence and average error after convergence plotted in log scale against percentage of measurements dropped for the simulated dataset.}
    \label{fig:measurement_loss}
\end{figure*}

The effects of measurement loss on the particle filter's convergence were investigated using the simulated dataset. A percentage of measurements to drop was varied from 0\% (all measurements present) to 100\% (no measurements present). The particle filter was run using the HMM motion model and the time to convergence and RMSE after convergence were recorded. The measurement loss percentage was applied evenly throughout the dataset, i.e. at 50\% data loss, every other measurement was dropped. Figure \ref{fig:measurement_loss} shows the results of this investigation.

Below 20\% measurement loss, the time to convergence is low, with an average of 43.8~s and a maximum of 129~s. Between 20\% and 80\% measurement loss, the average increases to 71.4~s and the maximum increases to 249~s. Above 80\% measurement loss, time to convergence increases drastically, with the average being 1342~s and the max being 6785~s. RMSE after convergence exhibits similar behavior: below 20\% measurement loss, the average RMSE is 15.07~m and the max is 23.18~m. Between 20\% and 80\% measurement loss, the average RMSE increases to 37.68~m and the max increases to 61.24~m. Above 80\% measurement loss, the average RMSE increases drastically to 391.61~m and the max increases to 1835.55~m.

This shows that the system is robust with up to 20\% of detections being lost, and can operate with up to 80\% of detections being lost with increased convergence times and RMSE.

\subsection{Depth Corrections} \label{depth_correction}

The hydrophones measure the 3D distance from the hydrophone to the shark, whereas the particle filter assumes that this distance is the 2D distance along the hydrophone's horizontal plane, which could cause errors in the transmitter's estimated position if the shark that the transmitter is attached to dives deep. Across the Shark Lab's white shark trajectory dataset, the maximum diving distance for any white shark was 10.7 meters. Due to the majority of shark tracking happening near the shorelines where biologists can more easily retrieve data and conduct experiments, the particle filter was developed and tested assuming that the depth of the transmitter is negligible. Assuming that the shark's depth stays constant at 10.7 meters, the true 2D distance is
\begin{align}
    r_{2D}(r_{meas}) = \sqrt{r_{meas}^2-10.7^2}.
\end{align}
where $r_{meas}$ is the measured distance returned by the hydrophone, and $r_{2D}$ is the true 2D distance required by the particle filter. The percentage error in the measurement can then be expressed as
\begin{align}
    \text{\% error} = \frac{r_{meas}-r_{2D}}{r_{2D}}.
\end{align}
When the hydrophone is near the transmitter, the percentage error is large, but it drops quickly as the hydrophone moves away. At 20 m, the percentage error is 18.4\%, at 50 m it is 2.4\%, and at 100 m it is 0.6\%. This supports the assumption that the depth is negligible for our dataset using the 2D assumption.

However, it is known that white sharks regularly dive to 400~m, and can on occasion reach depths of up to 1000~m \cite{Nasby2009}. If the shark were to dive deep relative to the distance to the hydrophone, the error would increase greatly. Although not presented in this paper, previous work has used transmitters that encode their depth in their signals, which enables the true distance in 3D to be calculated from the measured distance \cite{Lin2016}.

\section{Conclusions} \label{conclusions}
The results show that the particle filter is accurate to approximately 10~m, depending on the geometry of the hydrophones and transmitter. The best results were obtained with the HMM motion model fit to historical shark movement data and a 90 degree angle between the transmitter and hydrophones. The HMM motion model does best for the more realistic deployments where the transmitter moves substantially and is able to maintain multi-modal distributions when the transmitter is equally likely to be in one of two locations, although it can take longer to converge. The velocity motion model can also maintain these multi-modal distributions, although it is more likely to converge prematurely to the wrong possible transmitter location than the HMM motion model as shown by the simulated dataset results. The results of the measurement loss testing show that the system is robust with up to 20\% of detections being lost, and can operate with increased convergence times up to 80\% measurement loss.

The ability to conduct fish transmitter localization off-line, after tracking experiments have been conducted, has been shown to be successful. However, the system has additional use as a state estimation system that can be run online and act as an input to the AUV's planning and control system as shown in previous work \cite{Lin2016,Lin2013}. This would provide AUVs equipped with this newer, lower drag profile sensing system the ability to conduct real time autonomous tracking as done in previous work. As in previous work, an AUV-to-AUV communication system will be needed.

While the system in place has a variety of potential applications, the possibility of extending it to marine-life tracking to better improve conservation efforts has proven extremely valuable. The compactness of the AUV and hydrophone combination opens new doors to speed and maneuvering capabilities and, by extension, a stealthier method of tracking which puts less stress on both the system and the surrounding environment. Further, these new capabilities allow for longer and more accurate missions which will provide experts with a better understanding of wildlife behaviors and movement patterns. This knowledge will then assist with the identification of areas most suited for and requiring protection.

\section*{Acknowledgments}
This material is based on work supported by the National
Science Foundation under Grant No. 1952616. Field experiments detailed in this work were performed at Alamitos Bay and Santa Elena Bay with assistance from CSU Long Beach Shark Lab and the University of Costa Rica. White shark tracking data were funded by the State of California’s Shark Beach Safety Program. Preliminary tests of the system were conducted at the Robert J. Bernard Biological Field Station and Axelrood Aquatics Center at Roberts Pavilion. Work was also supported by UCR-VI C1127. Field logistics in Costa Rica were possible thanks to the support from Diving Center Cuajiniquil and Snorkeling Cuajiniquil. Field permits in Costa Rica were possible thanks to the Área de Conservación Guanacaste (Nº R-SINAC-ACG-PI-023-2022).

\bibliography{references}
\bibliographystyle{IEEEtran}

\end{document}